\documentclass[10pt,twocolumn,letterpaper]{article}

\usepackage{cvpr}
\usepackage{times}
\usepackage{epsfig}
\usepackage{graphicx}
\usepackage{amsmath}
\usepackage{amssymb}
\usepackage{subfigure}
\usepackage{dblfloatfix}

\usepackage{booktabs}
\usepackage{multirow}
\usepackage[normalem]{ulem}

\usepackage[pagebackref=true,breaklinks=true,letterpaper=true,colorlinks,bookmarks=false]{hyperref}

\cvprfinalcopy 

\def\cvprPaperID{51} 

\ifcvprfinal\fi
\begin{document}

\title{LATCH: Learned Arrangements of Three Patch Codes}

\author{Gil Levi and Tal Hassner\\
Department of Mathematics and Computer Science\\
The Open University of Israel\\
{\tt\small gil.levi100@gmail.com~~~~~~~~hassner@openu.ac.il}
}


\maketitle

\begin{abstract}
We present a novel means of describing local image appearances using binary strings. Binary descriptors have drawn increasing interest in recent years due to their speed and low memory footprint. A known shortcoming of these representations is their inferior performance compared to larger, histogram based descriptors such as the SIFT. Our goal is to close this performance gap while maintaining the benefits attributed to binary representations. To this end we propose the Learned Arrangements of Three Patch Codes descriptors, or LATCH. Our key observation is that existing binary descriptors are at an increased risk from noise and local appearance variations. This, as they compare the values of pixel pairs; changes to either of the pixels can easily lead to changes in descriptor values, hence damaging its performance. In order to provide more robustness, we instead propose a novel means of comparing pixel patches. This ostensibly small change, requires a substantial redesign of the descriptors themselves and how they are produced. Our resulting LATCH representation is rigorously compared to state-of-the-art binary descriptors and shown to provide far better performance for similar computation and space requirements.  
\end{abstract}

\section{Introduction}
The ability to effectively represent local visual information is key to a very wide range of computer vision applications. These applications range from image alignment, which requires that local image descriptors be accurately matched between different views of the same scene, to image classification and retrieval, where massive descriptor collections are frequently scanned in order to locate the ones most relevant to those of a query image. Consequently, computer vision research has devoted substantial efforts to develop and fine-tune these representations. 

\begin{figure}[t]
	\begin{center}
		\includegraphics[width=\linewidth]{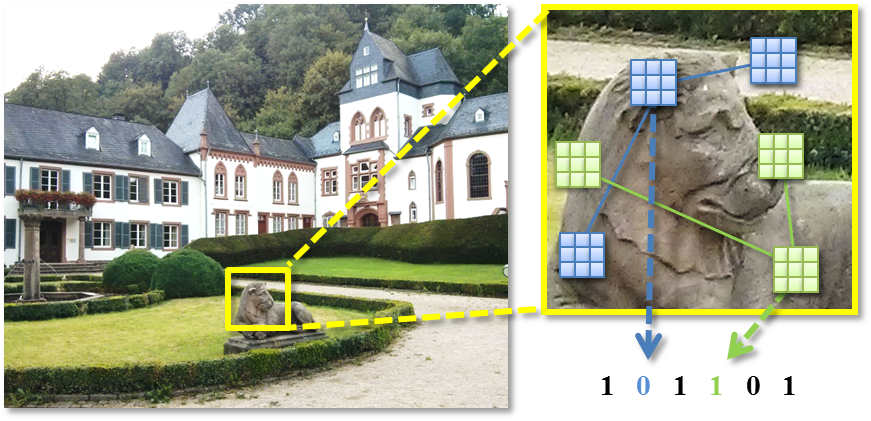}
	\end{center}
	\caption{{\bf Visualization of the LATCH descriptor.} Given an image patch centered around a keypoint, LATCH compares the intensity of three pixel patches in order to produce a single bit in the final binary string representing the patch. Example triplets are drawn over the patch in green and blue.\vspace{-5mm}}
	\label{fig:overview}
\end{figure}

At the core of the problem is the challenge of extracting local representations at keypoints, typically distributed sparsely over an image, in a manner which is both discriminative and invariant to various image transformations. Additional requirements, often as important if not more, are that a representation be efficient, in terms of the computational costs required to produce it, the space required to store it, and the time required to search for matching descriptors in large descriptor repositories. 

Over the past two decades, several distinct approaches for designing such descriptors have emerged. Two noteworthy designs are the distribution-based representations and the binary descriptors. Distribution-based descriptors, which include the successful SIFT~\cite{lowe2004distinctive} and HOG~\cite{dalal2005histograms} representations, represent visual information using distributions of image measurements (e.g., gradients, gradient orientations, etc.). Though proven highly effective in an ever widening range of applications, their main drawbacks are their size, the time required to produce them, and the challenges associated with efficiently searching through large numbers of such descriptors~\cite{heinly2012comparative}.

Binary descriptors, on the other hand, were designed with an emphasis on minimizing computational and storage costs~\cite{alahi2012freak,calonder2010brief,lepetit2013boosting,leutenegger2011brisk,rublee2011orb,strecha2012ldahash,trzcinski2013learning,trzcinski2012efficient}. These methods represent image patches using a (typically short) binary string, commonly computed by sampling and comparing pixels in the patch; different methods advocating different sampling strategies or other methods for increasing the descriptors discriminate power (e.g. boosting, discriminant projections). Though binary representations may not be as descriptive as their histogram counterparts, they make up for this shortcoming in their compact size, efficient computation, and the ability to quickly compare descriptor pairs using few processor-level instructions.

The representation presented here belongs to the latter family of descriptors. Our work is motivated by the longstanding observation that the act of sampling pixel pairs in order to compute each binary value in the representation is sensitive to noise and other changes in local appearances. Previous representations have addressed this problem by offering a number of alternative smoothing operations, which should be performed before the pixel values are sampled. Though this alleviated some of the problem, the unfortunate side effect of smoothing is, of course, the loss of information. This is particularly crucial in high-frequency regions of the image -- precisely where key points are detected, and where these representations are applied.

We offer an alternative approach based on the simple notion of comparing pixel patches rather than individual pixel values (illustrated in Fig.~\ref{fig:overview}). By comparing patches, visual information with more spatial support is considered for each of the descriptor's bits, and their values are therefore less sensitive to noise. We describe a patch-triplet based approach, in which triplets of patches are compared in order to set the binary values of the representation. Informative triplet arrangements are learned beforehand using labeled training data. Thus, triplet arrangements are ordered by their contribution to the successful classification of patches as being either similar or not while refraining from selecting highly correlated triplets. The most effective arrangements of patch triplets are then used to sample and compare patches whenever the descriptor is computed. 

The resulting representation, appropriately dubbed LATCH (Learned Arrangements of Three patCH codes), is evaluated extensively and shown to outperform existing alternatives by a wide margin, at the cost of a minor increase in the run-time computational requirements of extracting the descriptor. To summarize, this paper makes the following contributions.
\begin{itemize}
\item We propose a novel binary descriptor design, intended to provide improved stability and robustness than existing related descriptors. 
\item We show how effective descriptors can be generated by off-line, supervised learning of discriminative patch arrangements.
\item Extensive quantitative results and qualitative applications compare the capabilities of our LATCH representation with existing descriptors. These show LATCH to outperform other representations of its kind, significantly narrowing the performance gap between binary descriptors and histogram based methods. 
\end{itemize}

In order to promote reproducibility, as well as to allow easy use of our LATCH descriptors, our implementation is available online as part of the OpenCV library. Please see~\url{www.openu.ac.il/home/hassner/code.html} for further details.

\section{Related Work}\label{sec:related}
The development of local image descriptors has been the subject of immense research, and a comprehensive review of related methods is beyond the scope of this work. For a recent survey and evaluation of alternative binary interest point descriptors, we refer the reader to~\cite{heinly2012comparative}. Here, we only briefly review these and other related representations. \\

\noindent{\bf Binary descriptors.} Binary key-point descriptors were recently introduced in answer to the rapidly expanding sizes of image data sets and the pressing need for compact representations which can be efficiently matched. One of the first of this family of descriptors was the Binary Robust Independent Elementary Features (BRIEF)~\cite{calonder2010brief}. BRIEF is based on intensity comparisons of random pixel pairs in a patch centered around a detected image key point. These comparisons result in binary strings that can be matched very quickly using a simple XOR operation. As BRIEF is based on intensity comparisons, instead of image gradient computations and histogram pooling of values, it is much faster to extract than SIFT-like descriptors~\cite{lowe2004distinctive}. Furthermore, by using no more than 512 bits, a single BRIEF descriptor requires far less memory than its floating point alternatives.

Building upon BRIEF's design and matching method, the Oriented fast and Rotated BRIEF (ORB) descriptor~\cite{rublee2011orb} adds rotation invariance by estimating a patch orientation based on local first order moments within the patch. Another innovation proposed by~\cite{rublee2011orb} is the use of a unsupervised learning in order to select pixel pairs, rather than the random sampling of BRIEF. 

Rather than random sampling or unsupervised learning of pairs, the Binary Robust Invariant Scalable Keypoints (BRISK)~\cite{leutenegger2011brisk} use hand-crafted, concentric ring-based sampling patterns. BRISK uses pixel pairs with large distances between them to compute the patch orientation, and pixel pairs separated by short distances to compute the values of the descriptor itself, again, by performing binary intensity comparisons on pixel pairs. More recently, inspired by the retinal patterns of the human eye, the Fast REtinA Keypoint descriptor (FREAK) was proposed. Similarly to BRISK, FREAK also uses a concentric rings arrangement, but unlike it, FREAK samples exponentially more points in the inner rings. Of all the possible pairs which may be sampled under these guidelines, FREAK, following ORB, uses unsupervised learning to choose an optimal set of point pairs.

Similar to BRIEF, the Local Difference Binary (LDB) descriptor was proposed in~\cite{yang2012ldb,yang2014ldb} where instead of comparing smoothed intensities, mean intensities in grids of $2\times2$, $3\times3$ or $4\times4$ were compared. Also, in addition to the mean intensity values, LDB also compares the mean values of horizontal and vertical derivatives, amounting to 3 bits per comparison. Building upon LDB, the Accelerated-KAZE (A-KAZE) descriptor was suggested in~\cite{Alcantarilla13bmvc} where in addition to presenting a feature detector, the authors also suggest the Modified Local Difference Binary (M-LDB) descriptor. M-LDB uses the A-KAZE detector estimation of orientation for rotating the LDB grid to achieve rotation invariance and uses the A-KAZE detector's estimation of feature scale to sub-sample the grid in steps that are a function of the feature scale.   

A somewhat different descriptor design approach was proposed by~\cite{strecha2012ldahash}. Their LDA-Hash representation extracts SIFT descriptors from the image, projects them to a more discriminant space and then thresholds the projected descriptors to obtain binary representations. Though the final representation is a binary descriptor, producing it requires extracting SIFT descriptors, making the representation slower than its pure binary alternatives. To alleviate some of this computational cost, the DBRIEF~\cite{trzcinski2012efficient} representation projects patch intensities directly. The projections are further computed as a linear combination of a small number of simple filters from a given dictionary. Finally, the BinBoost representation of~\cite{lepetit2013boosting,trzcinski2013learning} also learns a set of hash functions that correspond to each bit in the final descriptor. Hash functions are learned using boosting and implemented as a sign operation on a linear combination of non linear week classifiers which are gradient based image features.

These last three representations, LDA-Hash, DBRIEF, and BinBoost, all obtain binary representations following application of filter combinations or floating-point descriptor extraction. Thus, though they claim improved performance over the original binary descriptors, they are all substantially more expensive computationally and so may therefore be unsuitable in many practical applications. 

Unlike these methods, our own uses efficient patch comparisons directly. Unlike the earlier representations (i.e. BRIEF,ORB,BRISK and FREAK), rather than comparing pairs of pixels, we compare {\em triplets of pixel patches} thereby providing more spatial support for each comparison. This provides more information at each comparison, making the binary values more robust to various sources of noise. Doing so also requires redesigning the descriptor itself. Finally, in contrast to the unsupervised learning of arrangements proposed by ORB, we use supervised learning to obtain efficient patch combinations. \\

\noindent{\bf Local binary patterns.} In a separate line of work, the Local Binary Patterns (LBP) were proposed as global (whole image) representation by~\cite{LBP2,ojala2002multiresolution}. Since their original release, they have been successfully applied to many image classification problems, most notably of texture and face images (e.g.,~\cite{ahonen2006face} and~\cite{nanni2012survey}). 

LBP produces for each pixel in the image a (typically very short) binary string representation. In fact, to our knowledge, in all reports of the use of LBP 8-bit strings or less were employed. These bits, similarly to the binary descriptors, are set following binary comparisons between image pixel intensities. In the original LBP implementation, these bits were computed by using a pixel's value as a threshold, applied to its eight immediate spatial neighbors, and taking the resulting zero/one values as the 8-bit string. By using only 8-bits, each pixel is thus represented by a code in the range of $[0..255]$ (or less, in some LBP variations), which are then pooled spatially in a histogram in order to represent image portions or entire images.

Our work is related to a particular LBP variant, the Three-Patch LBP (TPLBP)~\cite{wolf2011effective,wolf2008descriptor}, which was shown to be an exceptionally potent global representation for face images~\cite{guillaumin2009you}. Unlike previous LBP code schemes, TPLBP computes 8-bit value codes by comparing not the intensities of pixel pairs, but rather the similarity of three pixel patches. Specifically, for every pixel in the image, TPLBP compares the pixel patch centered on the pixel, with eight pixel patches, evenly distributed on a ring at radius $r$ around the pixel. A single binary value is set following a comparison of the center patch to two patches, spaced $\alpha$ degrees away from each other along the circle. A value of 1 represents the central patch being closer (in the SSD sense) to the first of these two patches, 0 otherwise. 

The TPLBP codes, though similar in spirit to the LATCH descriptor presented here, are different from it in several important aspects. Technically, the TPLBP uses a hand tailored, parameter controlled, limited sampling scheme, where a single anchor patch (the central patch) is compared again and again with a limited number of patch pairs at specific relative positions controlled by the ring radius $r$ and the angle between patches $\alpha$. Our proposed LATCH, on the other hand, can potentially consider {\em any} arrangement of three patches for this purpose. Moreover, LATCH {\em learns} which arrangements are optimal from training data, rather than being hand-crafted. 

More important, however, is the conceptual difference: LATCH is designed as a (sparse) key-point descriptor, rather than a per-pixel code intended for pooling over entire image regions. As far as we know, no previous work has considered using the design insights of the TPLBP to represent key points.

\section{Method}
\label{sec:Method}
We begin with a review of binary descriptor design. Let ${W}$ be a detection window, an image portion of fixed, pre-determined size, centered on a detected image key point. A binary descriptor $\mathbf{b}_W$ is formed by considering an ordered set $S=\{\mathbf{s}_t\}_{t=1\dots T}=\{[\mathbf{p}_{t,1},\mathbf{p}_{t,2}]\}_{t=1\dots T}$ of $T$ pairs of sampling coordinates, $\mathbf{p}_{t,1}=(x_{t,1},y_{t,1})$ and $\mathbf{p}_{t,2}=(x_{t,1},y_{t,2})$, given in ${W}$'s coordinate frame. The selection of values for $S$ is performed beforehand, either randomly (e.g., BRIEF~\cite{calonder2010brief}), manually (BRISK~\cite{leutenegger2011brisk}), or is automatically learned from training data (ORB~\cite{rublee2011orb} and FREAK~\cite{alahi2012freak}). 

Each index $t$ is typically associated not only with a pair of coordinates in ${W}$, but also with a pair of Gaussian smoothing kernels, $\mathbf{\sigma}_t = (\sigma_{t,1},\sigma_{t,2})_{t=1\dots T}$. These are applied separately to ${W}$, in the pixel coordinates given by $\mathbf{s}_t$, before being sampled. Thus, for each sampling pair $\mathbf{s}_t$, the smoothed intensities at the two sampling points $\mathbf{p}_{t,1}$ and $\mathbf{p}_{t,2}$, are compared and a single bit is set according to: 

\begin{equation} 
f(W,\mathbf{s}_t,\mathbf{\sigma}_t) = \begin{cases} 1 & \mbox{if } W(\mathbf{p}_{t,1},\sigma_{t,1}) > W(\mathbf{p}_{t,2},\sigma_{t,2}) \\ 0 & \mbox{otherwise} \end{cases} \label{eq:IntensityComparison}
\end{equation} 	

\noindent where $W(\mathbf{p}_{t,1},\sigma_{t,1})$ (similarly $W(\mathbf{p}_{t,2},\sigma_{t,2})$) is the value of the image window $W$ at coordinates $\mathbf{p}_{t,1}$ ($\mathbf{p}_{t,2}$) smoothed by a Gaussian filter with standard deviation $\sigma_{t,1}$ ($\sigma_{t,2}$). The final binary string $\mathbf{b}_W$, produced for image window $W$, is defined by 
\begin{equation} 
\mathbf{b}_W=\sum_{1\ \leq \ t \ \leq \ T}2^t f(W,\mathbf{s}_t,\mathbf{\sigma_t})  \label{eq:EqBuildingDescriptor} 
\end {equation}

\begin{table}[t]
\footnotesize{
	\centering
		\begin{tabular} {l @{~~~~~~~}  c  }
		\toprule
		Descriptor      & Running time (ms)\\ \hline
		SIFT~\cite{lowe2004distinctive} & 3.29\\ 
		SURF~\cite{bay2006surf} & 2.11 \\	
		LDA-HASH~\cite{strecha2012ldahash} & 5.03 \\ 
		LDA-DIF~\cite{strecha2012ldahash} & 4.74 \\ 
		DBRIEF~\cite{trzcinski2012efficient} & 8.75 \\ 
		BinBoost~\cite{lepetit2013boosting,trzcinski2013learning} & 3.29 \\\hline		
		BRIEF~\cite{calonder2010brief} & 0.234 \\
		ORB~\cite{rublee2011orb} & 0.486 \\
		BRISK~\cite{leutenegger2011brisk} & 0.059\\
		FREAK~\cite{alahi2012freak} & 0.072 \\
		A-KAZE~\cite{Alcantarilla13bmvc} & 0.069 \\ \hline
		LATCH & 0.616 \\
		\bottomrule
\end{tabular}
		\caption{{\bf Run time analysis}. Time measured in milliseconds for extracting a single local patch descriptor. Notice that LATCH only slightly slower than some of the popular binary descriptors and is an order of magnitude faster than the slower histogram and learning-based representations.}
		\label{tab:RunningTimes}
		}
\end{table}

\subsection{From pixel pairs to patch triplets}\label{sec:patchtriplets}
As previously mentioned, the pixel pairs sampling strategy presented above, though efficient, can be susceptible to noise as each bit relies on the values of two specific pixels. Though pre-smoothing can alleviate some of this problem, it can also result in the loss of information particularly at high frequency regions where key points are often detected. As a means of ameliorating this, we propose comparing pixel patches rather than pixels. Doing so, however, requires changing how each bit's value is set and in particular, defining a binary relation between pixel patches. This is achieved by using three-way patch comparisons. 

Specifically, we consider $t=1\dots T$ pixel patch {\em triplets}, adding the location of an ``anchor'' patch and redefining $S$ as $\hat{S}=\{\hat{\mathbf{s}}_t\}_{t=1\dots T}=\{[\mathbf{p}_{t,a},\mathbf{p}_{t,1},\mathbf{p}_{t,2}]\}_{t=1\dots T}$. Each of the pixel coordinates, $\mathbf{p}_{t,a}$, $\mathbf{p}_{t,1}$, and $\mathbf{p}_{t,2}$ provides the location of the central pixel in patches of size $k\times k$ pixels, denoted by $\mathbf{P}_{t,a}$, for the anchor patch, and $\mathbf{P}_{t,1}$, and $\mathbf{P}_{t,2}$ for its ``companion'' patches. We then evaluate the similarity of the anchor patch $\mathbf{P}_{t,a}$ to its two companions, by computing their Frobenious norm. Thus, the single binary value is produced by revising function $f$ as follows:

\begin{equation} 
g(W,\hat{\mathbf{s}}_t) = \begin{cases} 1 & \mbox{if } | ||\mathbf{P}_{t,a} - \mathbf{P}_{t,1}||^2_F > ||\mathbf{P}_{t,a} - \mathbf{P}_{t,2}||^2_F
\\ 0 & \mbox{otherwise} \end{cases} 
\end{equation} 	


\subsection{Learning patch triplet arrangements}
\label{subsec:LearningTheTriplets}
Even small detection windows $W$ give rise to a huge number of possible triplet arrangements. Considering that only a small number $T$ of bits is typically required (in practice, no more than 256), we must therefore consider which of the many possible triplet arrangements should be employed. Here, rather than taking one of the three approaches described by existing binary descriptors (Sec.~\ref{sec:related}), we propose our own selection criteria.

Specifically, we use the data-set introduced in~\cite{brown2011discriminative}. It consists of three separate collections: Liberty, Notre Dame, and Yosemite. Each of these contains over 400k local image windows that were extracted around multi-scale Harris corner detections~\cite{harris1988combined}. Pairs of these windows, extracted from different images in each collection, were labeled as being ``same'' (the two windows present the same physical scene point, viewed from different viewpoints or viewing conditions) or ``not-same''. These labels were obtained by employing multi-view stereo to form correspondences between different images in each collection. These windows were then partitioned into 500k pairs of which half are labeled as same and half not-same. 

We form 56k patch triplet arrangements, by random selection of the pixel coordinates $\mathbf{p}_{t,a}$ of the anchor patch, and the coordinates $\mathbf{p}_{t,1}$ and $\mathbf{p}_{t,2}$ of its two companion patches ($t=1\dots T=56,000$). We then evaluate each of these $T$ arrangements over all the window pairs in the benchmark, giving us 500k bits per arrangement. We define the quality of an arrangement by summing the number of times it correctly yielded the same binary value for ``same'' labeled pairs and different values for ``not-same'' labeled pairs. 

Arrangement selection based on this criteria may result in highly correlated arrangements being selected. To prevent this, following~\cite{alahi2012freak,rublee2011orb}, we add arrangements incrementally, skipping over those with responses highly correlated to previously selected arrangements. Specifically, a candidate arrangement is selected if its absolute correlation with all previously selected arrangements is smaller than a threshold $\tau$. In our experiments, this value was set to $\tau=0.2$ and left unchanged.

We note that others have also used the data-set from~\cite{brown2011discriminative} for the purpose of learning binary descriptors (e.g.~\cite{lepetit2013boosting,strecha2012ldahash,trzcinski2013learning,trzcinski2012efficient}). However, those methods differ from the one proposed here as they do not learn optimal arrangements for pixel comparison, but instead learn optimal projections or linear/non-linear filters to apply to these patches. The method presented here is simpler, yet provides comparative, or even better performance, as we later show.


\begin{table*}[t!]
\footnotesize{
	\centering
		\begin{tabular} { l  c  c  c  c  c  c  c  c  c}
		\toprule
		Descriptor &  Bark & Bikes &  Boat  & Graffiti & Leuven& Trees & UBC  & Wall & Average \\ \hline
		SIFT~\cite{lowe2004distinctive}  & 0.077 & 0.322 &0.080 &0.127 &0.130 &0.047 &0.130 &0.138 &0.131 \\ 
		SURF~\cite{bay2006surf}		& 0.071 &0.413 &0.088 &0.133 &0.300 &0.046 &0.268 &0.121 &0.180\\ 	
		LDA-HASH~\cite{strecha2012ldahash} & 0.199 & 0.466 & 0.269 & 0.155 & 0.303 & 0.110 & 0.393 & 0.268 & 0.270\\
		LDA-DIF~\cite{strecha2012ldahash} & 0.197 & 0.472 & 0.278 &0.170 &0.435 &0.101 & 0.396 & 0.260 & 0.289\\
		DBRIEF~\cite{trzcinski2012efficient} & 0.000 & 0.025 & 0.001 &0.008 &0.010 &0.001 &0.031 &0.002 &0.010\\
		BinBoost~\cite{lepetit2013boosting,trzcinski2013learning} & 0.055 &0.344 &0.083 &0.132 &0.338 &0.037 &0.217 &0.119 & 0.166\\	\hline
		BRIEF~\cite{calonder2010brief} & 0.055 & 0.353 & 0.050 & 0.102 & 0.227 & 0.060 & 0.178 & 0.141 & 0.146 \\ 
		ORB~\cite{rublee2011orb} & 0.032 & 0.208 & 0.048 & 0.062 & 0.118 & 0.027 & 0.121 & 0.050 & 0.083\\ 
		BRISK~\cite{leutenegger2011brisk} & 0.015 & 0.138 & 0.026 & 0.071 & 0.161 & 0.018 &0.131 & 0.038 & 0.075 \\
		FREAK~\cite{alahi2012freak} & 0.019 & 0.145 & 0.034 & 0.101 & 0.194 & 0.026 &0.147 & 0.041 & 0.089\\ 
		A-KAZE~\cite{Alcantarilla13bmvc} & 0.022 & 0.326 & 0.005 & 0.048 & 0.138 & 0.027 & 0.144 & 0.048 &0.095 \\\hline
		LATCH &0.065 &0.415 &0.057 &0.119 &0.374 &0.082 &0.215 &0.175 &0.188\\ 
		\bottomrule
		\end{tabular}
		\caption{{\bf Oxford benchmark results.} Numerical results summarizing area under the recall vs. 1-precision curve for the eight subsets of the Oxford set. Results for the much larger, floating point, histogram based descriptors are presented separately. Higher results are better.  In total, LATCH outperforms almost all alternatives, including even the floating point descriptors such as SIFT and SURF.\vspace{-5mm}}
		\label{tab:MikolajczykResults}
		}
\end{table*}

\begin{figure}[t!]
\centering{
				\includegraphics[width=0.40\textwidth,clip,trim = 30mm 10mm 30mm 10mm]{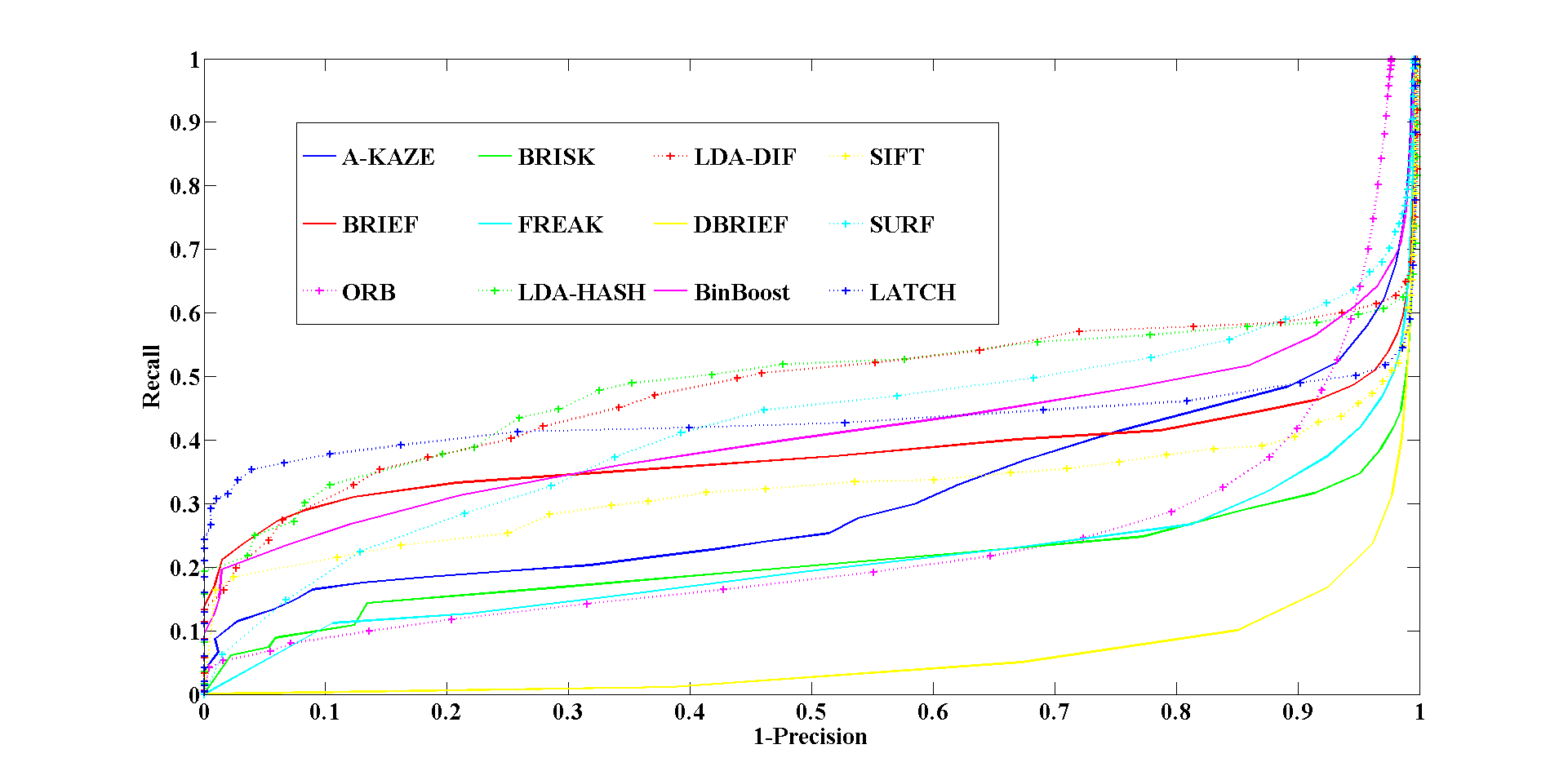}\\
        \includegraphics[width=0.40\textwidth,clip,trim = 30mm 10mm 30mm 10mm]{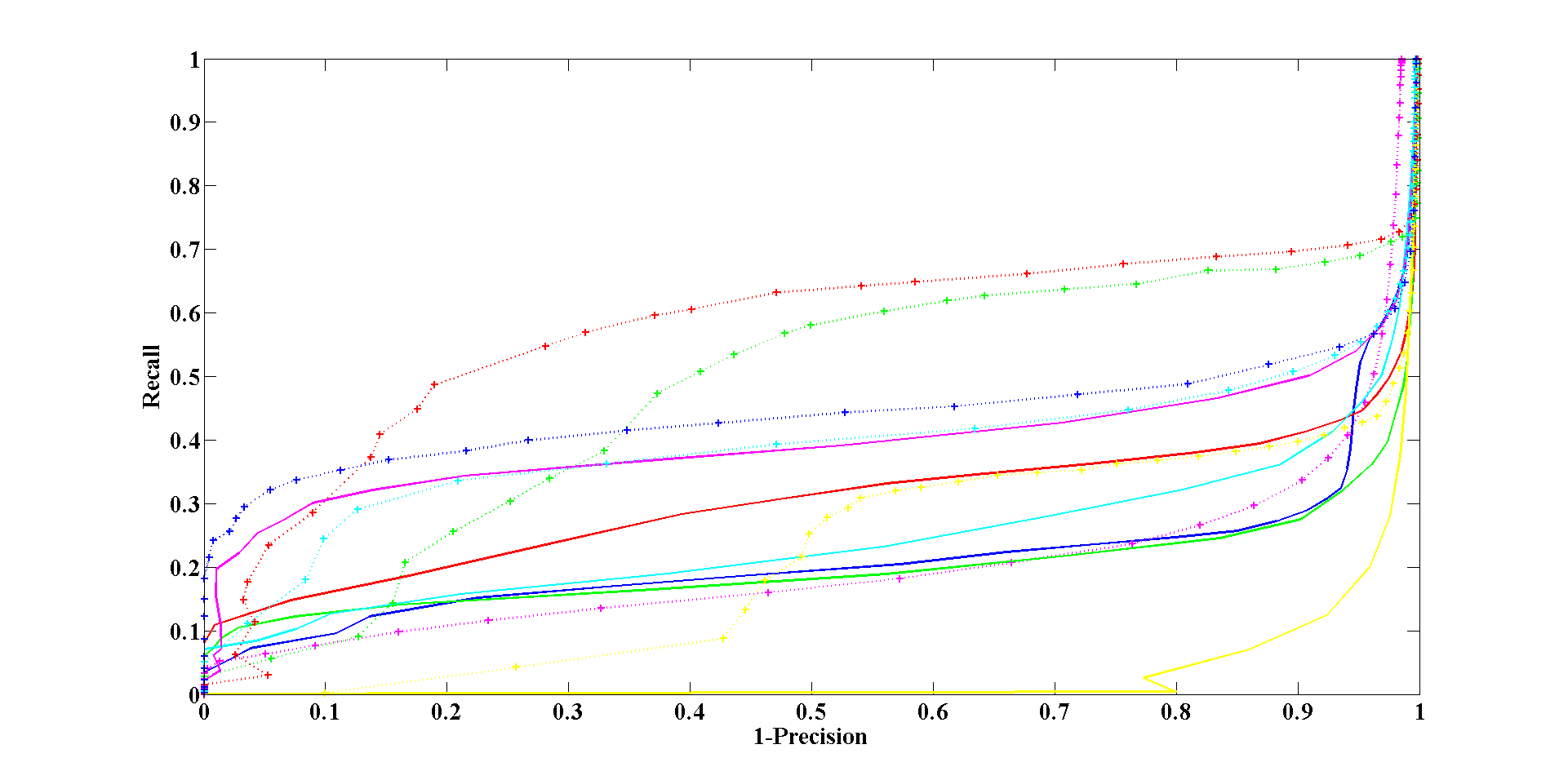}        
				}
        \caption{{\bf Oxford benchmark Recall vs. 1-precision curves.} Top: Bikes results (blur) ; Bottom: Leuven results (lightning). Evidently, LATCH outperforms all methods, except those that are an order of a magnitude slower. Notice LATCH's superior performance at the high precision section of the graph.\vspace{-5mm}}
				\label{fig:MikolajczykGraphs}
\end{figure}

\section{Experimental results}
\label{sec:Experiments}
Our LATCH extraction routine is implemented in C$++$ using OpenCV 2.0 for image processing operations. Unless otherwise noted, we used 32-byte LATCH descriptors with $7\times 7$ patches. Detection windows are $48\times 48$ pixels centered on key-points. Our tests use the efficient C$++$ descriptor implementations available from OpenCV or from their various authors, with parameter values left unchanged. 

\subsection{Empirical results}\label{sec:empirical}

Comparisons are provided using a wide range of relevant alternative methods. These include the ``pure''-binary descriptors: BRIEF~\cite{calonder2010brief}, ORB~\cite{rublee2011orb}, BRISK~\cite{leutenegger2011brisk}, FREAK~\cite{alahi2012freak} and A-KAZE~\cite{Alcantarilla13bmvc}. We additionally provide results comparing LATCH to the more computationally expensive LDA-Hash~\cite{strecha2012ldahash}, DBRIEF~\cite{trzcinski2012efficient} and BinBoost~\cite{lepetit2013boosting,trzcinski2013learning} representations. Finally, the performances of SIFT~\cite{lowe2004distinctive} and SURF~\cite{bay2006surf} are also provided.

We used two standard benchmarks for our tests: the Oxford~\cite{mikolajczyk2005performance,mikolajczyk2005comparison} and the Learning Local Image Descriptors~\cite{brown2011discriminative} benchmarks. Our tests employ the test protocols associated with these benchmarks. We additionally provide a range of tests designed to evaluate the contribution and effect of various design aspects of our LATCH descriptor.\\

\noindent {\bf Run times.} We begin by comparing the computational costs associated with extracting the various descriptors used in our experiments. The time (ms) required to extract a single descriptor were averaged over 250K patches of different scale and orientation, taken from various images. Measurements were performed on an Intel Core i7 laptop with 16.0 GB of memory, running 64-bit Microsoft Windows 8.1. 

Table~\ref{tab:RunningTimes} summarizes the measured running times. The substantial difference between the time required to extract the pure binary descriptors, including our own LATCH, and descriptors based on floating point values is clearly evident. In particular, LATCH requires an order of magnitude less time than some of these alternatives.\\

\noindent{\bf Oxford data-set.} Originally described by~\cite{mikolajczyk2005performance,mikolajczyk2005comparison} this set has since become the standard for evaluating descriptor design capabilities, and in particular, the capabilities of the binary descriptors discussed here (see, e.g.,~\cite{alahi2012freak,Alcantarilla13bmvc,calonder2010brief,leutenegger2011brisk}).

The Oxford data-set comprises of eight image sets, each with six images presenting increasing appearance variations. The appearance variations modeled by the benchmark sets are: zoom and rotation (the Boat and Bark sets), planar perspective transformations (view-point changes in the Graffiti and Wall sets), lightning changes (the Leuven set), JPEG compression (the UBC set), and increasing degrees of blur (the sets Bikes and Trees).

For each set, we compare the first image against each of the remaining five and check for correspondences. Performance is measured using the code from~\cite{mikolajczyk2005performance,mikolajczyk2005comparison}\footnote{Available from:~\url{www.robots.ox.ac.uk/~vgg/research/affine}}, which computes recall and 1-precision using known ground truth homographies between the images. We also provide the area under the recall vs. 1-precision curve, averaged over all five image pairs in each set.

Following the test protocol employed in, e.g.,~\cite{alahi2012freak,Alcantarilla13bmvc,bay2006surf,calonder2010brief,leutenegger2011brisk,rublee2011orb} each descriptor was extracted at image locations detected using its own original key-point detector. Our own LATCH descriptor was applied to key-points returned by the multi-scale Harris based detector used by the original SIFT implementation~\cite{lowe2004distinctive}. As some of the sets in the Oxford benchmark depict rotation changes and some do not, we implement rotation invariance by using the detected orientation, or the descriptors' own estimates when available.

Table~\ref{tab:MikolajczykResults} summarizes our results. Fig.~\ref{fig:MikolajczykGraphs} additionally provides recall vs. 1-precision curves for the datasets Bikes and Leauven. Aside from LDA-HASH and LDA-DIF which extract binary descriptors by first extracting SIFT descriptors and thus are much slower, LATCH outperforms the other binary descriptors on most of the sets and in some cases even the much larger, histogram representations, SIFT and SURF. \\

\noindent {\bf Learning Local Image Descriptors data-set.} We next report tests on the data-set described by~\cite{brown2011discriminative}\footnote{Available from:~\url{www.cs.ubc.ca/~mbrown/patchdata/patchdata.html}}. It provides a large number of detection windows along with same/not-same labels signifying whether two windows from two separate images correspond to the same physical point or not.

The test protocol used here is designed to evaluate the discriminative power of different image descriptors. Given two windows, a descriptor is extracted for each one and the distance between the two descriptors is measured. A scalar threshold is then applied to this distance in order to determine if the two descriptors are similar enough to imply that the windows should be labeled ``same'' or not. We use the Yosemite dataset in order to learn an optimal threshold by using linear support vector machines (SVM)~\cite{svm}. Yosemite images were also used to learn patch triplet arrangements for the LATCH descriptor (Section~\ref{subsec:LearningTheTriplets}). Testing is performed on the Liberty and Notre-Dame sets.

\begin{figure*}[h!]
\begin{tabular}{c@{}c@{}c@{}c}
\includegraphics[width=0.24\textwidth,clip,trim = 30mm 0mm 30mm 0mm]{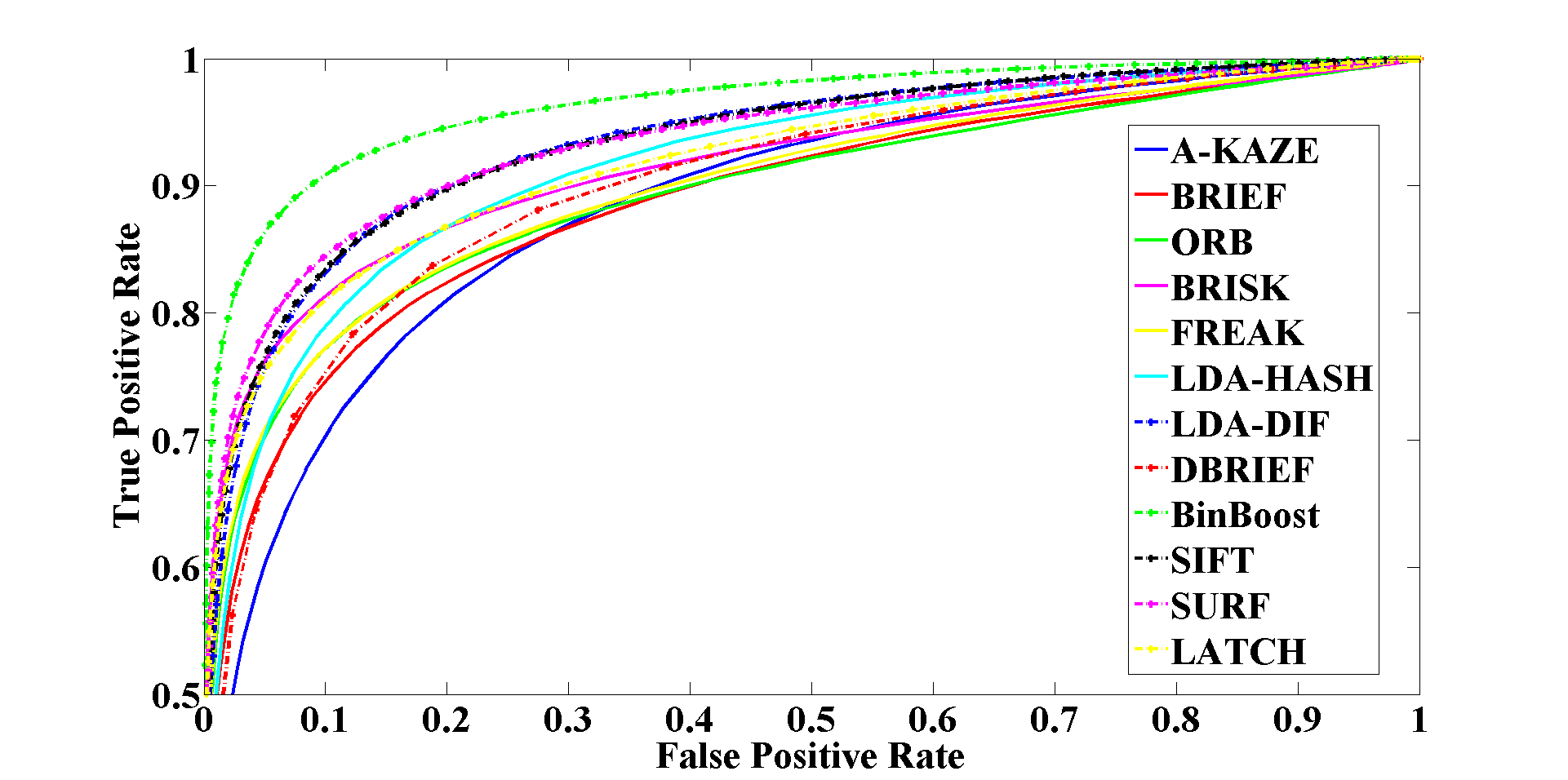}&
\includegraphics[width=0.24\textwidth,clip,trim = 30mm 0mm 30mm 0mm]{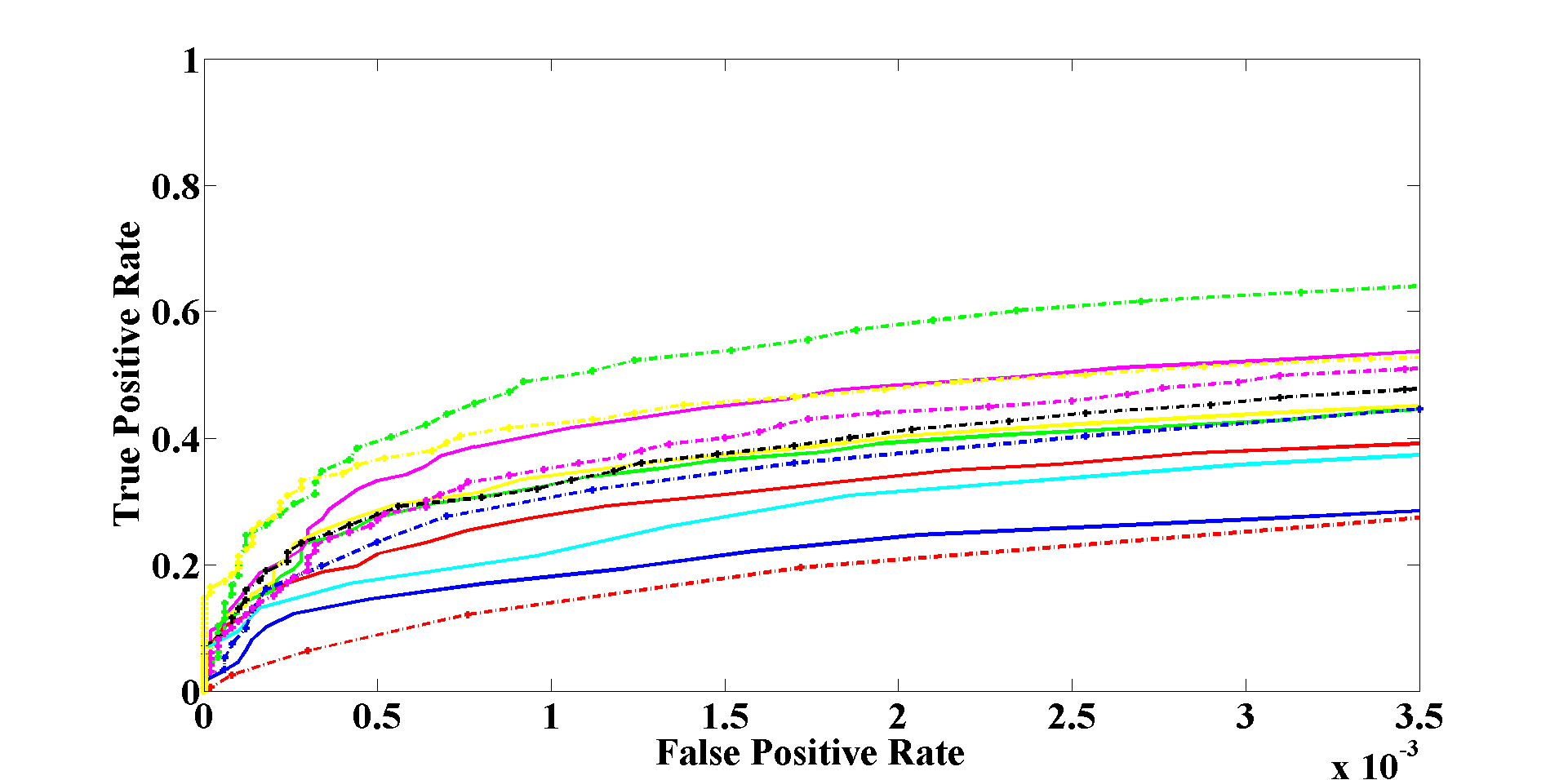}&
\includegraphics[width=0.24\textwidth,clip,trim = 30mm 0mm 30mm 0mm]{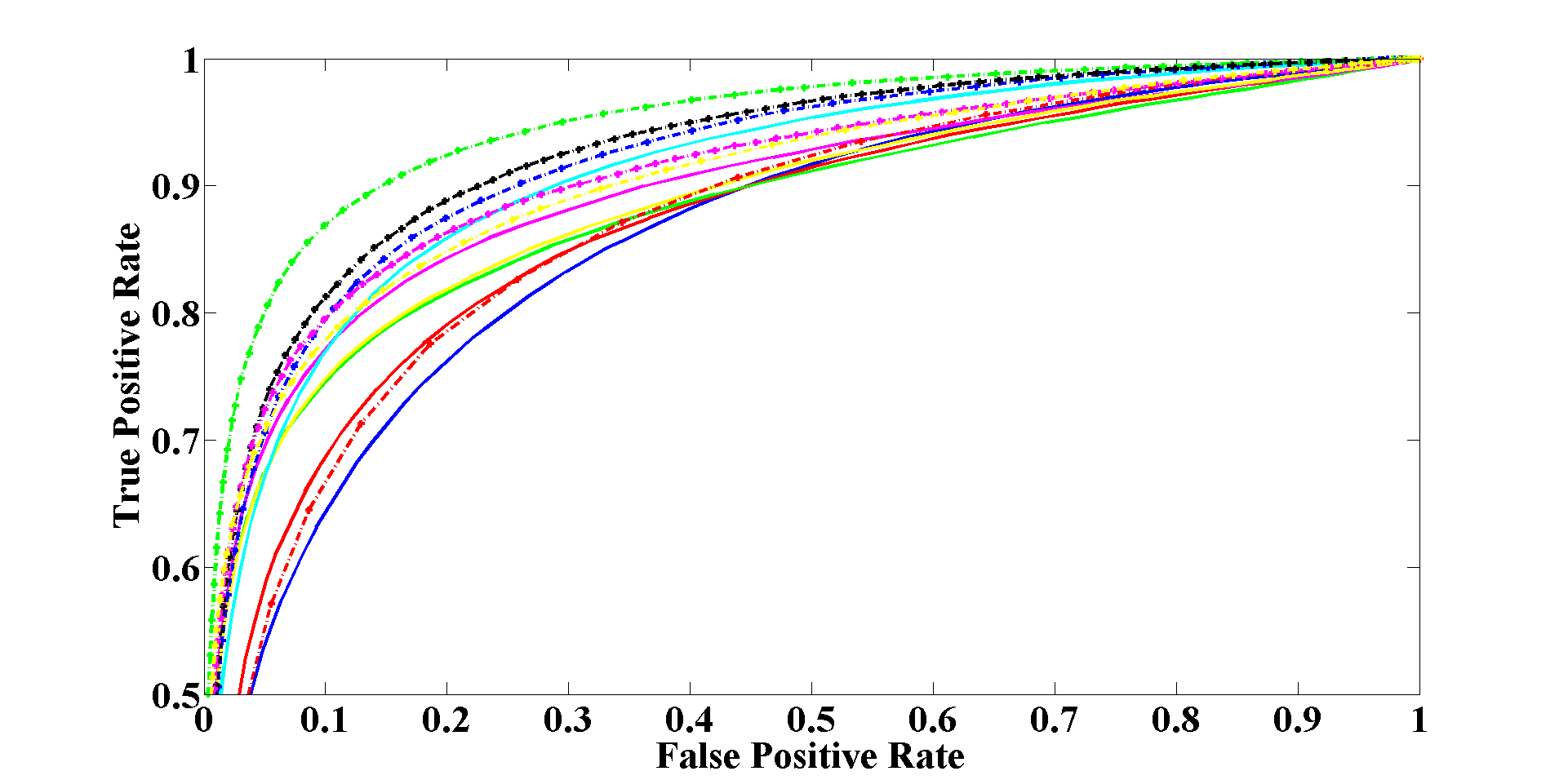}&
\includegraphics[width=0.24\textwidth,clip,trim = 30mm 0mm 30mm 0mm]{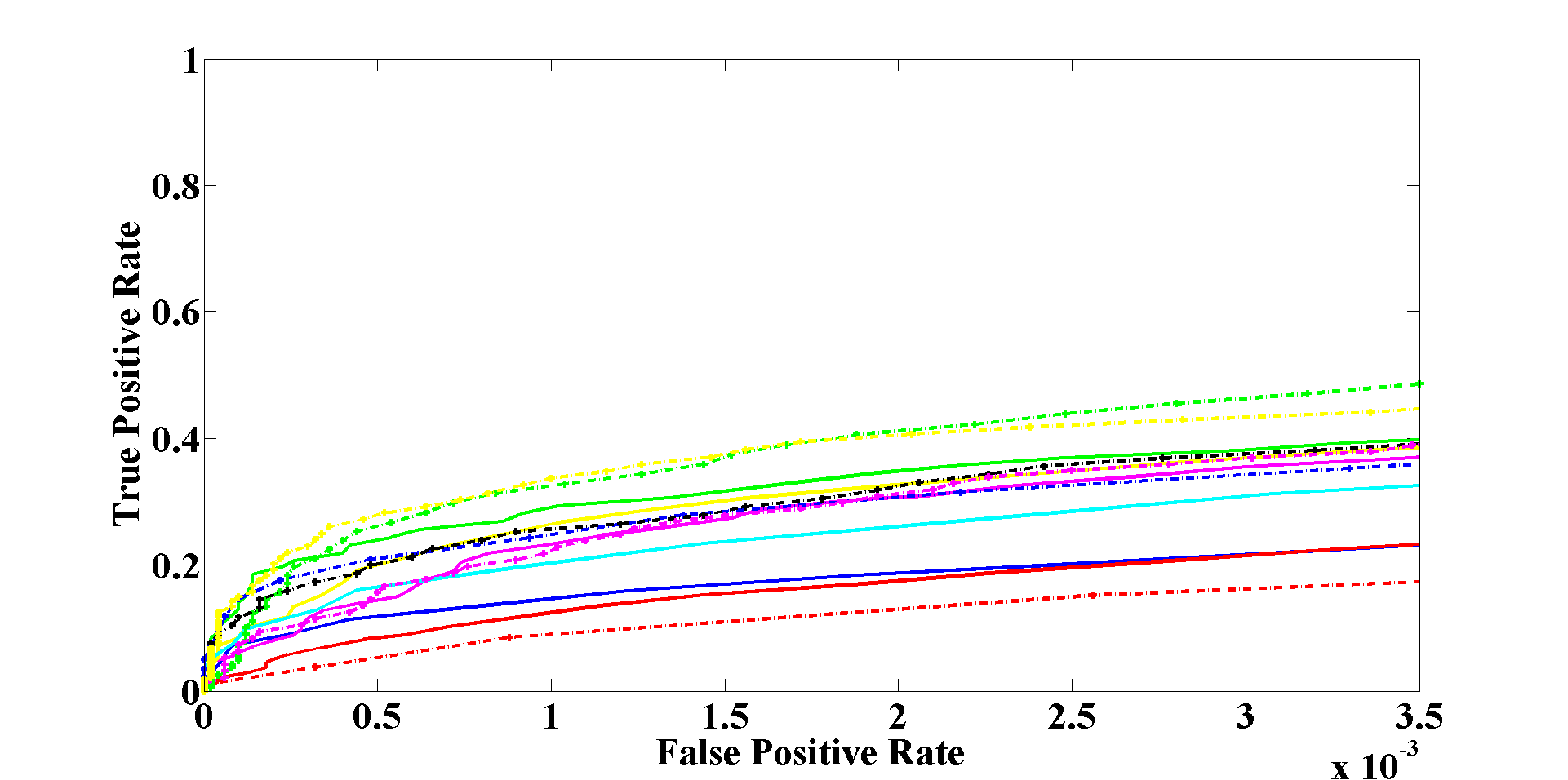}\\
(a) & (b) & (c) & (d)
\end{tabular}				
        \caption{{\bf ROC curves for the Learning Local Descriptors data-set tests.} (a) Notre Dame set, ROC curves; (b) zoomed-in view of the low false positive region of the ROC for the Notre-Dame tests (c) ROC curves for the Liberty tests; (d) zoomed-in view of the low false positive region of the ROC for the Liberty tests. High resolution versions of these figures are available in the supplemental material.}
				\label{fig:SameNotSameGraphs}
\end{figure*}

Table~\ref{tab:sameNotSameResults} summarizes the results in terms of accuracy, area under the ROC curve and 95\% error-rate (the percent of incorrect matches obtained when 95\% of the true matches are found). ROC curves for the different methods tested are presented in Fig~\ref{fig:SameNotSameGraphs}. Our results show the clear advantage of the proposed LATCH descriptor over other binary descriptor designs, with LATCH outperforming the other representations, in both tests, by noticeable margins. Although BinBoost and LDA-HASH/DIF perform better than LATCH on these tests, as previously noted, this added performance comes at substantial computational costs.\\

\noindent {\bf Analysis: Varying descriptor size.} In the tests reported above, we used a LATCH descriptor of 32 bytes. Here, we revisit the tests on the Oxford benchmark in order to evaluate the effect descriptor size has on its performance. We test varying descriptor sizes (the number of arrangements used) using 4, 8, 16, 32, and 64 bytes for the representation. Table~\ref{tab:analysis} (a) summarizes our results, providing the area under the recall vs. 1-precision curve. Clearly, the performance of LATCH improves as its size grows. These results can be compared with those of Table~\ref{tab:MikolajczykResults}. \\

\begin{table}[th]
\footnotesize{
\centering
\begin{tabular} {l @{~~} c@{~~}c@{~~}c c@{~~}c@{~~}c}
\toprule
& \multicolumn{3}{c}{Notre-Dame} & \multicolumn{3}{c}{Liberty}\\
Descriptor  & AUC & ACC & 95\% Err & AUC & ACC & 95\% Err\\
\hline
SIFT~\cite{lowe2004distinctive} & .934 & .817 &39.7 &.928 & .764 &40.1 \\
SURF~\cite{bay2006surf}		&.935 &.866 &41.1 &.911 &.833 & 55.0\\ 
LDA-HASH~\cite{strecha2012ldahash} &.916 &.830 &46.7 &.910 &.798 &48.1 \\
LDA-DIF~\cite{strecha2012ldahash} &.934 &.857 &38.5 &.921 &.836 &43.1 \\
DBRIEF~\cite{trzcinski2012efficient} &.900 &.830 &55.1 &.868 &.794 &61.5 \\
BinBoost~\cite{lepetit2013boosting,trzcinski2013learning} &.963 &.907 &21.6 &.949 &.884 &29.3 \\ \hline
BRIEF~\cite{calonder2010brief} & .889 & .823 &63.2  & .868 & .798 & 66.7 \\
ORB~\cite{rublee2011orb}   &.894 &.835 & 66.2& .882 & .822 & 69.2 \\
BRISK~\cite{leutenegger2011brisk} &.915 &.857 &57.7 &.897  &.834 &62.6 \\ 
FREAK~\cite{alahi2012freak} &.899 &.835 &61.5 &.887 &.824 &65.0 \\
A-KAZE~\cite{Alcantarilla13bmvc} & .885 & .806 & 56.7 & .860  & .782  & 63.4 \\ \hline
LATCH & .919 &.855 &52.0 &.906  &.838 &56.7  \\
\bottomrule
\end{tabular}
\caption{{\bf Results on the Learning Local Descriptors dataset.} Same/not-same tests on data from~\cite{brown2011discriminative}. Testing was performed separately on the Notre-Dame and the Liberty collections. Higher results are better for AUC and accuracy (ACC); lower results are better for the 95\% error-rate (Err.). Evidently, LATCH outperforms other binary representations by clear margins.}
\label{tab:sameNotSameResults}
}
\end{table}


\begin{table*}[t]
\footnotesize{
	\centering
		\begin{tabular} {l l @{~~~~~~~} c c c c c c c c c}
		\toprule
		& Descriptor      &  Bark & Bikes &  Boat  & Graffiti & Leuven& Trees & UBC  & Wall & AVG  \\ \hline
		\multirow{5}{*}{(a) Descriptor size}  
		& LATCH-4 &0.012 &0.282 &0.013 &0.038 &0.269 &0.014 &0.131 &0.030 &0.099 \\
		& LATCH-8 &0.029 &0.353 &0.028 &0.074 &0.319 &0.039 &0.167 &0.080 &0.136\\
		& LATCH-16 &0.053 &0.394 &0.044 &0.098 &0.355 &0.064 &0.192 &0.133 &0.167\\
		& LATCH-32 &0.065 &0.415 &0.057 &0.119 &0.374 &0.082 &0.215 &0.175 &0.188\\
		& LATCH-64 &0.073 &0.425 &0.070 &0.131 &0.381 &0.097 &0.239 &0.205 &0.203\\ 
		\hline
		\multirow{8}{*}{(b) Patch size}  
		& LATCH $1\times 1$ &0.058 &0.391 &0.048 &0.103 &0.346 &0.069 &0.190 &0.139 &0.168 \\
		& LATCH $3\times 3$  &0.054 &0.392 &0.049 &0.105 &0.361 &0.070 &0.193 &0.133 &0.170\\ 
		& LATCH $5\times 5$ &0.064 &0.405 &0.054 &0.113 &0.368 &0.076 &0.205 &0.156 &0.180\\ 
		& LATCH $7\times 7$ &0.065 &0.415 &0.057 &0.119 &0.374 &0.082 &0.215 &0.175 &0.188\\ 
		& LATCH $9\times 9$ &0.072 &0.422 &0.059 &0.123 &0.374 &0.085 &0.221 &0.188 &0.193\\
		& LATCH $11\times 11$ &0.075 &0.428 &0.058 &0.128 &0.376 &0.085 &0.223 &0.196 &0.196\\
		& LATCH $13\times 13$ &0.078 &0.429 &0.057 &0.129 &0.372 &0.085 &0.220 &0.200 &0.196\\
		& LATCH $15\times 15$ &0.074 &0.434 &0.054 &0.126 &0.367 &0.081 &0.216 &0.200 &0.194\\		
		\hline
		\multirow{4}{*}{(c) Learning method}  
		& Random &0.064 &0.391 &0.059 &0.104 &0.260 &0.064 &0.229 &0.174 &0.168 \\
		& ORB/FREAK &0.073 &0.396 &0.066 &0.108 &0.267 &0.074 &0.239 &0.187 &0.176 \\
		& Proposed &0.055 &0.413 &0.058 &0.107 &0.379 &0.083 &0.229 &0.155 &0.185 \\
		& Combined &0.065 &0.415 &0.057 &0.119 &0.374 &0.082 &0.215 &0.175 &0.188 \\ 		
		\bottomrule
\end{tabular}
		\caption{{\bf Analysis tests on the Oxford benchmark}. Results summarizing the performance of our LATCH descriptor using (a) different descriptor sizes (different numbers of patch triplet arrangements); (b) different patch sizes; (c) different methods of selecting arrangements. The table provides area under the recall vs. 1-precision curves. Please see text for more details.\vspace{-3mm}}
		\label{tab:analysis}
		}
\end{table*}

\begin{table}[!th]
\footnotesize{
\centering
\begin{tabular} {l @{~~}c @{~~}c @{~~}c c@{~~} c @{~~}c}
\toprule
& \multicolumn{3}{c}{Notre-Dame} & \multicolumn{3}{c}{Liberty}\\
Learning Method  & AUC & ACC & 95\% Err. & AUC & ACC & 95\% Err.\\
\hline
Random & .894 & .822 & 57.5  & .871 & .793 & 60.6 \\
ORB/FREAK & .902 & .831 & 55.4 & .881 & .801 & 58.2 \\
Proposed & .905 & .842 & 58.6 & .892  &  .824 & 61.8 \\ 
Combined & .919 & .855 & 52.0 & .906 & .838 &56.7 \\
\bottomrule
\end{tabular}
\caption{{\bf Analysis of different learning methods on the Learning Local Descriptors dataset.} Same/not-same tests with different learning methods performed using the data from~\cite{brown2011discriminative}. Testing was performed separately on the Notre-Dame and the Liberty collections. Higher results are better for AUC and accuracy (ACC); lower results are better for the 95\% error-rate (Err.). Interestingly, our proposed learning method outperforms~\cite{alahi2012freak,rublee2011orb}. When combining their correlated triplet elimination technique (``combined'') we gain a further performance boost.\vspace{-05mm}}
\label{tab:LearningMethodBrown}}
\end{table}

\noindent {\bf Analysis: Varying patch size.} One of the key components of the LATCH descriptor is the use of pixel patches compared to sampling single pixels. We next evaluate the effect of larger pixel patches on the performance of LATCH. Here, we use a 32 byte LATCH representation, testing it with patches of sizes $3\times 3$, $5\times 5$, $7\times 7$, $9\times 9$, $11\times 11$, $13\times 13$ and $15\times 15$. 

We report also the performance of a simpler LATCH variant, which is computed by comparing {\em pixel} triplets, rather than patch triplets (LATCH $1\times 1$). Similarly to ORB, in order to handle noise pixel values are sampled following the same local smoothing. Extracting larger-patch LATCH descriptors following smoothing brought performance further down, and so we do not report these results.

Our results, summarized in Table~\ref{tab:analysis} (b) demonstrate that larger patches provide more accuracy. In nearly all cases, the bigger the patches used, the higher the performance gain. The relative improvement in performance, however, decays with patches larger than $9\times 9$ or $11\times 11$. Here too, these results may be compared with those presented in Table~\ref{tab:MikolajczykResults}, where LATCH was computed using $7\times 7$ patches.

It is worthwhile to consider the performance of LATCH $1\times 1$. Evidently, this approach almost always provides inferior results even to LATCH extracted using $3\times 3$ patches. With the default $7\times 7$ patches, LATCH performance is significantly better than sampling single pixels.\\

\begin{figure*}[t!]
\centering{
				\includegraphics[width=\linewidth,clip,trim = 1mm 3mm 1mm 3mm]{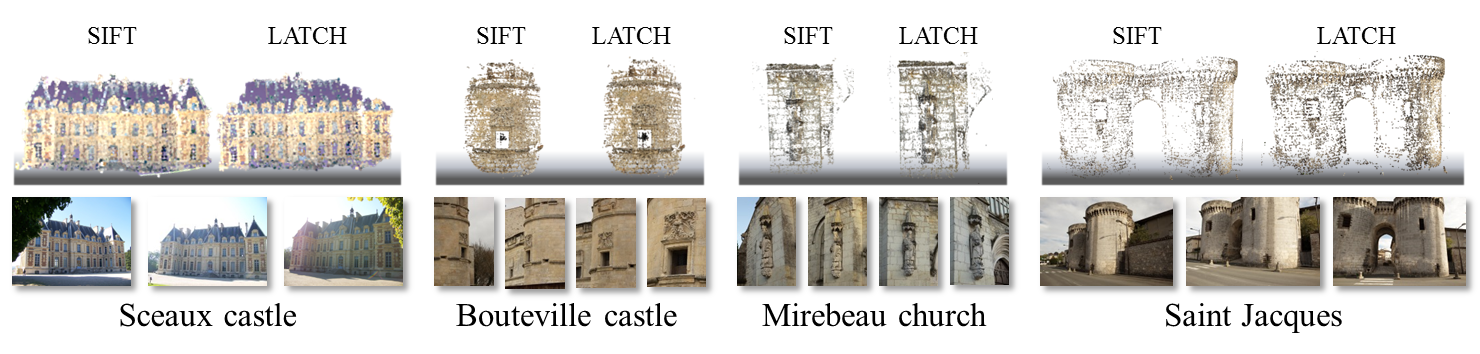}\vspace{-3mm}
				}
        \caption{{\bf Structure from motion results.} Top: 3D reconstruction results on four standard test sequences obtained with the incremental structure from motion chain method~\cite{moulon2013adaptive} using the default SIFT and our LATCH. Bottom: Input image examples from each set. Qualitatively, SIFT and LATCH provide comparable results though LATCH descriptor matching is an order of magnitude faster (Table~\ref{tab:SFMruntime}).\vspace{-3mm}}
				\label{fig:SFM}
\end{figure*}

\noindent {\bf Analysis: Comparing learning methods.} As discussed in~\ref{subsec:LearningTheTriplets}, we propose supervised learning of optimal patch arrangements. The same approach can of course be applied to learning optimal pairs. We compare the proposed approach to that of ORB~\cite{alahi2012freak,rublee2011orb} and also present the performance of the combined method in which the quality of the triplets is measure by their score on the ``same''/``not-same'' dataset, filtering out  correlated triplets. 

Table~\ref{tab:analysis} (c) presents results on the Oxford benchamark and Table~\ref{tab:LearningMethodBrown} on the Learning Local Descriptors set. Evidently, overall, the proposed learning method outperforms the learning method of~\cite{alahi2012freak,rublee2011orb}, and the combined method outperforms both. Unsurprisingly, random selection of patch triplets performs much worse than either of these.

\subsection{Application to multi-view 3D reconstruction}\label{sec:mvs}
One of the more challenging uses of local descriptors lies in structure from motion (SfM) applications. In order to produce accurate results, local appearances must be matched across images of the same scene, taken from possibly widely different views. Additionally, SfM methods often compare many comparisons between many interest points, and hence the efficiency of matching descriptors is also a matter of concern.

We test the use of our proposed LATCH descriptor in a SfM framework, comparing it to the SIFT descriptor often used for this purpose. To this end, we have incorporated LATCH into the OpenMVG library~\cite{moulon2013adaptive} using their incremental structure from motion chain method. We ran SfM twice, changing only the local image representations from their default SIFT to our own LATCH descriptors. 

In order to isolate the effect of using LATCH rather than SIFT, both use the same key-points, recovered by the SIFT detector implemented in the VLFeat library~\cite{vedaldi08vlfeat}. All OpenMVG parameters were kept at their default values apart from the ratio threshold which was 0.6 for SIFT (the default), and raised to 0.99 for LATCH (binary descriptors in general are known to be more sensitive to this value).


\begin{table}[!th]
\footnotesize{
\centering
\begin{tabular} {l c c}
\toprule
Sequence & SIFT & LATCH\\
\hline
Sceaux Castle & 381.63 & 39.05\\
Bouteville Castle & 4766.22 & 488.70\\
Mirebeau Church & 3166.35 & 325.31\\
Saint Jacques & 1651.12 & 169.19\\
\bottomrule
\end{tabular}
\caption{{\bf Structure from motion descriptor matching times.} The time (seconds) required to match descriptors when producing the structure from motion results reported in Fig.~\ref{fig:SFM}. LATCH is consistently an order of magnitude faster to match than the standard SIFT, yet provides qualitatively similar results.}
\label{tab:SFMruntime}}
\end{table}

Reconstruction results for standard test image sequences~\cite{moulon2013adaptive} are provided in Fig.~\ref{fig:SFM} and the time required to match the descriptors in each scene is provided in Table~\ref{tab:SFMruntime}. We note that denser surfaces could conceivably be produced by running a multi-view stereo algorithm, e.g. the Patch-based Multi-View Stereo (PMVS) method of~\cite{furukawa2010accurate}, following the initial reconstructions. Doing so, however, may correct errors due mismatching descriptors. We focus on the quality of the descriptors, not the final reconstruction, and so this step was not performed here.

Evidently, 3D reconstructions obtained by using both descriptors are qualitatively comparable. The time required to match our LATCH descriptors, however, is consistently an order of magnitude faster than SIFT.

\section{Conclusions}
\label{sec:conclusions}
Over the years, the computer vision community has invested immense efforts in a continuing effort to improve the performance of local descriptors, including the requirements they make on storage, extraction and matching time. As part of this effort, we propose a new variant to the binary descriptors representation family. Our LATCH representation enjoys the same fast matching time and small storage requirements of binary descriptors. Our tests, however, demonstrate that it outperforms other binary descriptors by wide margins, closing the gap between their performance and the performance reported by the much larger, more expensive histogram based representations.

At the heart of our representation lies the observation that by sampling individual pixel pairs, previous descriptors may be at risk of being too sensitive to noise and other local changes in appearance. Instead, we consider comparisons of pixel patches. In order to provide meaningful comparisons, we propose comparing patch triplets, rather than pixel pairs, and provide a means for selecting triplets which provide the most discriminative capabilities. We test this representation extensively and provide a comprehensive comparison with other similar representations.


{\small
\bibliographystyle{ieee}
\bibliography{LATCH}
}

\end{document}